\documentclass{Interspeech}
\interspeechcameraready

\title{Dialogue Response Prefetching Based on Semantic Similarity and \\ Prediction Confidence of Language Model}

\author[affiliation={1,2}]{Kiyotada}{Mori}
\author[affiliation={2,1}]{Seiya}{Kawano}
\author[affiliation={2}]{Angel}{Fernando Garcia Contreras}
\author[affiliation={3,2,1}]{Koichiro}{Yoshino}

\affiliation{}{Nara Institute of Science and Technology}{Japan}
\affiliation{Guardian Robot Project}{RIKEN}{Japan}
\affiliation{}{Tokyo Institute of Technology}{Japan}

\email{mori.kiyotada.mh5@naist.ac.jp,
seiya.kawano@riken.jp,
angel.garciacontreras@riken.jp,
yoshino.k.ai@m.titech.ac.jp}

\usepackage{comment}

\begin{document}
\maketitle
\begin{abstract}
Prefetching of dialogue responses has been investigated to reduce user-perceived latency (UPL), which refers to the user's waiting time before receiving the system's response, in spoken dialogue systems. To reduce the UPL, it is necessary to predict complete user utterances before the end of the user's speech, typically by language models, to prepare prefetched dialogue responses. In this study, we proposed a prediction confidence model (PCM) that determines whether prefetching is possible or not by estimating the semantic similarity between the predicted complete user utterance and the complete user utterance. We evaluated our PCM based on the differences between the predicted complete user utterance and the complete user utterance.
\end{abstract}

\section{Introduction}
With the progress of research into speech recognition and dialogue response generation, cascade-type spoken dialogue systems (SDSs) are becoming more widely used. Cascade-based SDSs generate spoken responses following two steps \cite{ji2024wavchat}. An automatic speech recognition (ASR) model transcribes a user's speech. A dialogue model generates a response from the transcription. In this sequence, the SDSs generate the response after end-of-sentence/speech (EOS), which is determined by the voice activity detection (VAD) model of ASR. However, this framework causes response delays because the user must wait for the system to generate a response after the user's utterance has ended, which decreases the naturalness of an SDS \cite{jacoby2024human}.

Speech response prefetching, which prepares dialogue responses before a user has finished speaking, has been proposed to solve this problem \cite {Kamioka2024response, Zink2024predictive}. If prefetching is successful, the system can respond immediately after detecting EOS. One of the current response prefetching procedures assumes incremental speech recognition and uses a language model to predict complete user utterances from partially recognized utterances. The language model used to predict complete user utterances is called the prediction model. It addresses the delay problem by preparing the system's speech response in advance for the predicted complete user utterance. The system uses a language model to predict the confidence score for the predicted complete utterance to determine the prefetching timing. The language model implemented to determine prefetching timing is called a prediction confidence model (PCM).

Existing PCMs are modeled to use confidence scores for predicting the subsequent user utterance at the character or word level. 
However, when considering the actual system response generation, it is important to capture the intent of the user utterance and prepare an appropriate response in advance rather than perfectly predicting every word of the user utterance \cite{skantze2005exploring}. In other words, prefetching success depends on whether the user's intent is captured and an appropriate response is prepared.

In this study, we refined PCMs to use the semantic meaning of a user utterance. We trained the PCM using the semantic similarity between complete user utterances and predicted user utterances by a prediction model rather than estimating the literal matching between them.
In addition, we defined successful prefetching as the ability to generate a system response from the predicted user utterance that is of comparable quality to that generated from the complete user utterance, rather than merely considering the successful prediction of the user utterance, as done in existing research.
The operation of our concept is compared with that of existing work in Fig. 1. In the conventional PCM if the predicted complete utterance is "What is the weather" and the complete user utterance is "What is the weather $like$ $today?$", it is a negative example for training a PCM because the predicted user utterance does not contain the word "$like$ $today?$". On the other hand, it is a positive example for training our PCM because such utterances are considered semantically similar.

\begin{figure}[t]
\begin{minipage}[b]{1\columnwidth}
    \centering
    \includegraphics[width=0.75\columnwidth]{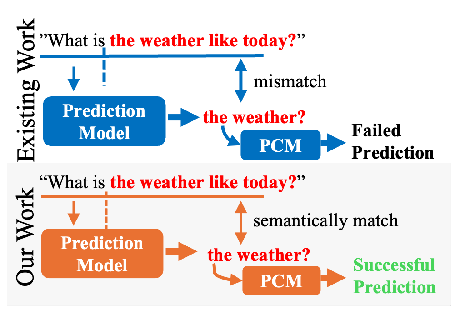}
    \caption[]{Comparison of proposed PCM with conventional PCM in related work}
    \label{fig:1}
\end{minipage}
\end{figure}

In constructing our PCM, we investigated the relationship between the predicted and complete user utterances in terms of semantic similarity, based on BERT scores and the success rate of prefetching. We compared our PCM with existing prefetching systems that operate on the word level matching between predicted and complete user utterances. We examined our framework using task-oriented dialogue (TOD) benchmarks: MultiWOZ \cite{zang2020multiwoz}, Spoken-MultiWOZ (SpokenWOZ) \cite{si2024spokenwoz}, and Japanese MultiWOZ (JMultiWOZ) \cite{ohashi2024jmultiwoz}.
These different benchmarks were used because PCMs may show different behaviors depending on language structure and the head position of the language. We experimented using gold transcriptions rather than the actual ASR hypothesis with errors, assuming that the speech recognition accuracy was sufficiently high. 

Our contributions are as follows.
\begin{enumerate}
\item We constructed a PCM based on the semantic similarity between predicted and complete user utterances.
\item We examined the proposed semantic PCM and confirmed that it could prefetch natural responses more quickly than existing word-level models.
\item We experimented on several TOD datasets and confirmed the differences between English and Japanese languages.
\end{enumerate}

\section{Preliminaries}
\subsection{Problem Definition}
Personalized Predictive ASR  \cite{schwarz2023personalized} proposed a pipeline method that prefetches a spoken response and reduces user-perceived latency (UPL). The definition of UPL is shown in Equation (1).
\begin{align}
T_{\mathrm{UPL}} = 
\begin{cases} 
    \mathrm{max} (T_{\mathrm{EP}},T_{\mathrm{PF}}+T_{\mathrm{Response}}) & (\text{Successful}) \\
    T_{\mathrm{EP}}+T_{\mathrm{Response}}  & (\text{Failed}).
\end{cases}
\end{align}
$T_{\mathrm{EP}}$ is the time for EOS detection. $T_{\mathrm{Response}}$ is the time for the system's response generation, and $T_{\mathrm{PF}}$ is the time for the system's response generation after the PCM decides to prefetch the response. The goal of dialogue response prefetching is to minimize $T_{\mathrm{UPL}}$ while maintaining the quality of prefetched responses. When the definition of any prefetching success is satisfied before the EOS by $\Delta T$, $T_{\mathrm{PF}}=-\Delta T$ is defined. $\Delta T$ is the time gain from this prefetching, which the previous authors called prediction gain. Minimizing $T_{\mathrm{UPL}}$ thus means maximizing the prediction gain.

The prediction model and the PCM are defined as:
\begin{align}
\hat{y}_{\mathrm{full},t} \approx \underset{y_{\mathrm{full}}}{\mathrm{argmax}}\  P(y_{\mathrm{full}}|\hat{y}_{t}),
\end{align}
\begin{align}
P(\hat{y}_{\mathrm{full},t}=\hat{y}_{\mathrm{full}}).
\end{align}
$\hat{y_t}$ is the partial user utterance at time $t$, while $y_{\mathrm{full}}$ and $\hat{y}_{\mathrm{full}}$ respectively indicate the complete user utterance and the final ASR hypothesis corresponding to the complete user utterance. $\hat{y}_{\mathrm{full},t}$ is the predicted utterance of $\hat{y_t}$ to $\hat{y}_{\mathrm{full}}$.  In this study, we assume that the speech recognition accuracy is sufficiently high so that $\hat{y}_{\mathrm{full}}=y_{\mathrm{full}}$.

In this framework, the system incrementally recognizes the user's speech and then predicts the complete user utterance from the recognized part using a prediction model. 
PCM is applied to the predicted user utterance to estimate the probability that the predicted user utterance and the complete user utterance match. The system response is prefetched if this probability exceeds the threshold before detecting EOS; otherwise, the system response is generated when EOS is detected.
\subsection{Look-ahead techniques}
The look-ahead technique  \cite{ohagi2024investigation} assumes that UPL can be reduced by preparing candidate user utterances that might appear in the dialogue context before the user speaks. By considering the measured semantic similarity of the user utterance candidates prepared in advance and the user utterances obtained from ASR, it became possible to reduce the time cost of decoding for dialogue response generation. This method predicted 20$\%$ of the user utterances in the Dialogue Robot Competition 2022 \cite{robotcomp2022}, where humans evaluated more than 80$\%$ of the prefetched system responses as natural.

If we refined a PCM's output based on semantic similarity rather than word-level matching, we could expect an improvement in prediction gain similar to this model's success.

\section{PCM Based on Semantic Similarity}
Since the existing PCM has a strict criterion of exact word-string matching, there was a problem where prefetching often failed. Therefore, we refined our PCM to estimate the probability that the semantic similarity between the predicted utterance and the complete utterance is greater than a threshold. The definition of our PCM is given in Equation (4).
\begin{align}
P\left(\mathrm{S\text{-}BERT}(\hat{y}_{\mathrm{full},t}, \hat{y}_{\mathrm{full}}) > T\right),
\end{align}
where S-BERT is the Sentence BERT (stsb-xlm-r-multilingual) \cite{reimers-2019-sentence-bert} and $T$ is a threshold.

Our PCM was constructed by following labels defined in the experiment to fine-tune the CLS vector of BERT (bert-base-multilingual-uncased)  \cite{DBLP}. We confirmed the following two points in the experiment for this redefinition.

\begin{enumerate}
    \item What is the optimal threshold $T$?
    \item Can our PCM prefetch responses that are comparable to responses to complete user utterances?
\end{enumerate}

\section{Experiment}
\subsection{Dataset}
MultiWOZ and JMultiWOZ were used to train and evaluate the PCMs for English and Japanese, respectively. SpokenWOZ was used to measure prediction gain from the time stamp of each word. The following shows the procedure for constructing the training data set. The difference between English and Japanese datasets is whether characters or words are used as the unit of time $t$.
\begin{enumerate}
    \item Separate each user utterance of the MultiWOZ into words.
    \item According to each word input from the beginning, a prediction model predicts a complete user utterance.
    \item For the complete user utterance, a language model ( the same model used as the prediction model in this experiment) generates four system responses.
\end{enumerate}

The English and Japanese prediction models were Qwen (Qwen2.5-14B-Instruct)  \cite{qwen2.5} trained on MultiWOZ and JMultiWOZ, respectively. These have been fine-tuned using Low-Rank Adaptation (LoRA)  \cite{hu2021lora} on the user utterances of validation datasets. The inputs are the belief state, up to 4 past dialogue utterances, partial utterances, and a response example. Utterances that do not have dialogue history are not used because they become difficult to predict. At this time, the hyperparameters are as follows: epochs are 1, the learning rate is $2e\text{-}4$, the batch size is 32, the LoRA rank is 16, the optimization function is Adam 8 bit, the maximum number of input tokens is 2048, and the temperature is 1.

The following elements were used in the training dataset for the PCM:
\begin{itemize}
\item $\hat{r}_{\mathrm{full},t}$: System response generated by the prediction model for $\hat{y}_{\mathrm{full},t}$ when $\hat{y}_{\mathrm{full}} \ne \hat{y}_t$
\item $\hat{r}_{\mathrm{full}}$: 4 system responses generated by the prediction model for $\hat{y}_{\mathrm{full}}$. 
\item $r_{\mathrm{full}}$: System response in the test dataset for $\hat{y}_{\mathrm{full}}$
\item $h_{\mathrm{dialogue}}$: Up to four past utterances before $\hat{y}_{\mathrm{full}}$
\end{itemize}
Four $\hat{r}_{\mathrm{full}}$ were used only for response evaluation with multiple references. Otherwise, one randomly sampled $\hat{r}_{\mathrm{full}}$ was used for evaluation, as explained below.
\subsection{Labels}
To train and evaluate the PCM, we generated two types of labels: $l_{\mathrm{sbert}T}$ and $l_{\mathrm{literal}}$. $l_{\mathrm{sbert}T}$ is a label that is positive when $\text{S-BERT}(\hat{y}_{\mathrm{full},t},\hat{y}_{\mathrm{full}})>T$, and negative otherwise. In this experiment, $T = \{0.75, 0.80, 0.85, 0.90, 0.95\}$. On the other hand, $l_{\mathrm{literal}}$ is a label that is positive when $y_{\mathrm{full},t} = y_{\mathrm{full}}$ and negative otherwise. This label is based on the same definition of 
successful prefetching used in the original PCM.
\subsection{Training}
The PCMs for English and Japanese were fine-tuned on the user utterances (50 dialogues) from the test datasets of MultiWOZ, SpokenWOZ, and
JMultiWOZ, respectively. The input features for the PCMs are $h_{\mathrm{dialogue}}$, $\hat{y}_{t}$, and $\hat{y}_{\mathrm{full},t}$. The hyperparameters used for training the PCM are as follows: epoch is $1$, the learning rate is $2e-5$, the batch size is $16$, the loss function is Focal Loss ($\gamma=2.0$)  \cite{ross2017focal}, and the output layer is the Softmax function.
\subsection{Evaluation}
We evaluated English and Japanese PCMs based on the user utterances in the test data sets, which were not used during training, of MultiWOZ/SpokenWOZ and
JMultiWOZ, respectively. The punctuation in $\hat{y}_{\mathrm{full},t}$ and $\hat{y}_{\mathrm{full}}$ of SpokenWOZ was removed because the writing style of SpokenWOZ did not follow written text, such as written fillers with commas or without commas.
To evaluate the PCMs, we had to consider three points: the success or failure of training, the prediction gain, and the naturalness of the prefetched system responses, as evaluated by both machines and humans. 
\subsubsection{Automatic Evaluation}
The success or failure of the PCM training was evaluated using the following three indicators.
\begin{itemize}
    \item Successful Prefetch Rate (SPR): The percentage of utterance in which the PCM determines that a prediction was successful for the first time, and the definition of successful prefetching is met
    \item Failed Prefetch Rate (FPR): The percentage of utterance in which the PCM determines that a prediction was successful for the first time, but the definition of successful prefetching is not met
    \item Non-Prefetch Rate (NPR): The percentage of utterances in which the PCM does not determine that a prediction was successful until the EOS
\end{itemize}
The prediction gain obtained by the PCM was evaluated using the following three indicators.
\begin{itemize}
    \item P-Gain ($\%$): The average proportion of utterances from EOS to prefetching, representing the prediction gain when prefetching is successful
    \item P-Gain (ms): The average utterance duration (time) from EOS to prefetching, representing the prediction gain when prefetching is successful; this was measured only from PCMs trained on SpokenWOZ
    \item C-Gain: Difference in number of characters between $\hat{y}_{t}$ and $\hat{y}_{\mathrm{full}}$ when prefetching is successful
\end{itemize}
The machine evaluation of the prefetched system response group was performed using the following five indicators. Athena-RR \cite{harrison2023transformer} was used as the response ranking model.
\begin{itemize}
    \item Total: Number of successful prefetching
    \item Comp: Percentage of $\hat{r}_{\mathrm{full}}  \ne  \hat{r}_{\mathrm{full},t}$ when prefetching is successful
    \item S-BERT: Maximum semantic similarity between $\hat{r}_{\mathrm{full},t}$ and the four $\hat{r}_{\mathrm{full, n}}$, defined as $\underset{n=1,2,3,4}{\mathrm{max}} \mathrm{S\text{-}BERT}(\hat{r}_{\mathrm{full},t}, \hat{r}_{\mathrm{full},n})$
    \item ROUGE: Maximum F1 score for ROUGE-1 between $\hat{r}_{\mathrm{full},t}$ and four $\hat{r}_{\mathrm{full, n}}$, defined as $\underset{n=1,2,3,4}{\mathrm{max}} \mathrm{ROUGE}(\hat{r}_{\mathrm{full},t}, \hat{r}_{\mathrm{full},n})$
    \item PR $<$ R: Proportion evaluations by Athena-RR where $\hat{r}_{\mathrm{full}}$ is evaluated more highly than $\hat{r}_{\mathrm{full},t}$, given $h_{\mathrm{dialogue}}$, $\hat{y}_{\mathrm{full}}$, and $r_{\mathrm{full}}$ as input when $\hat{r}_{\mathrm{full}} \ne \hat{r}_{\mathrm{full},t}$
\end{itemize}

Since Athena-RR only supports English, the input for evaluating the Japanese PCM was translated into English using GPT3.5 Turbo \cite{brown2020language}.

\subsubsection{Human Evaluation}
We confirmed that prefetched responses were not out of context in dialogue, an essential requirement for dialogue. We conducted the human evaluation of the English and Japanese responses in the prefetched system response group using the following procedure. For Japanese response evaluation, three native Japanese subjects were recruited. For English response evaluation, five subjects with English proficiency of CEFR B2 \cite{council2001common} or higher were recruited. Each Japanese and English sample was evaluated by three subjects. The experimental procedure is outlined below.
\begin{enumerate} 
\item 300 responses are randomly sampled from pairs of $\hat{r}_{\mathrm{full},t}$ and $\hat{r}_{\mathrm{full}}$. The two responses are shuffled so that the participants cannot distinguish which one is prefetched.
\item A participant evaluates $\hat{r}_{\mathrm{full},t}$ and $\hat{r}_{\mathrm{full}}$ in terms of naturalness on a 5-point scale (1: not natural at all, 2: not very natural, 3: neither good nor bad, 4: quite natural, 5: extremely natural). 
\item A participant compares the responses to determine which is more natural for $\mathrm{h}_{\mathrm{dialogue}}$ and $\hat{y}_{\mathrm{full}}$, allowing for a 3-point scale with options PR$<$R: $\hat{r}_{\mathrm{full},t} < \hat{r}_{\mathrm{full}}$, PR$>$R: $\hat{r}_{\mathrm{full},t} > \hat{r}_{\mathrm{full}}$, and PR$=$R: $\hat{r}_{\mathrm{full},t} = \hat{r}_{\mathrm{full}}$.
\end{enumerate}

The naturalness ratings were integrated using the average score, while the comparative evaluations were integrated based on the mode. If the comparative evaluations by the subjects were split among PR$<$R, PR$>$R, and PR$=$R, the result was standardized as PR$=$R for consistency. We did not show ${r}_{\mathrm{full}}$ to subjects because our response evaluation was not intended to compare human response quality with a language model's response quality but rather to confirm whether sufficient response quality could be maintained.

\begin{table*}[htbp]
\centering
\caption{Prediction and response evaluation of PCMs trained by MultiWOZ and JMultiWOZ}
\begin{tabular}{l|lll|ll|lllll}
    \hline
     Model& SPR&  FPR &  NPR &  P-Gain $(\%)$ $\uparrow$ &  C-Gain $\uparrow$ &  Total &  Comp &  ROUGE $\uparrow$ &  S-BERT $\uparrow$ &  PR$<$R \\
    \hline
   $\text{Multi}_{l_{\mathrm{sbert}075}}$&0.25&0.25&0.51&0.51&20.71&
  2761 &0.99&0.46&0.68&0.50\\
   $\text{Multi}_{l_{\mathrm{sbert}080}}$
   &0.25&0.24&0.51&0.51&20.71&2674&0.99&0.48&0.68&0.51\\
   $\text{Multi}_{l_{\mathrm{sbert}085}}$ &0.22&0.27&0.52&0.36&14.80 &2329 &0.98&0.50&0.69&0.50\\
   $\text{Multi}_{l_{\mathrm{sbert}090}}$ &0.19&0.29&0.52&0.28&11.39& 2029 &0.97&0.53&0.71&\textbf{0.49}\\
   $\text{Multi}_{l_{\mathrm{sbert}095}}$ &0.14&0.33&0.52&0.28&11.39& 1491 &0.97&0.55&0.73&\textbf{0.49}\\
   $\text{Multi}_{l_{\mathrm{literal}}}$ &0.09&0.35&0.56&0.12&4.29&
  892 &0.96&0.56&0.74&\textbf{0.48}\\
  \hline
   $\text{JMulti}_{l_{\mathrm{sbert}075}}$ &0.33&0.17&0.50&0.61&16.97&
  1051 &0.95&0.65&0.73&0.54\\
   $\text{JMulti}_{l_{\mathrm{sbert}080}}$ &0.31&0.18&0.51&0.58&15.71&1000&0.95&0.67&0.76&0.53\\
   $\text{JMulti}_{l_{\mathrm{sbert}085}}$ &0.29&0.20&0.51&0.53&13.97 &940 &0.95&0.68&0.77&0.52\\
   $\text{JMulti}_{l_{\mathrm{sbert}090}}$ &0.26&0.24&0.51&0.49&12.59& 815 &0.95&0.70&0.78&0.53\\
   $\text{JMulti}_{l_{\mathrm{sbert}095}}$ &0.20&0.29&0.50&0.47&11.47&657 &0.95&0.73&0.81&\textbf{0.49}\\
   $\text{JMulti}_{l_{\mathrm{literal}}}$ &0.19&0.30&0.51&0.18&4.40&
  582 &0.95&0.76&0.85&\textbf{0.49}\\
  \hline
\end{tabular}
\end{table*}
\begin{table*}[htbp]
\centering
\caption{Prediction and response evaluation of PCMs trained by SpokenWOZ}
\footnotesize
\begin{tabular}{l|lll|lll|lllll}
    \hline
     Model&  SPR&  FPR &  NPR&  P-Gain (\%) $\uparrow$&  P-Gain (ms) $\uparrow$&  C-Gain $\uparrow$&  Total&  Comp &  ROUGE $\uparrow$&  S-BERT $\uparrow$&  PR$<$R \\
    \hline
   $\text{Spoken}_{l_{\mathrm{sbert}075}}$ &0.26&0.19&0.55&0.32&1061.25&9.36&5677&1.00&0.40&0.57&0.50\\
     $\text{Spoken}_{l_{\mathrm{sbert}080}}$ &0.25&0.20&0.56&0.28&914.14&8.16&5265&1.00&0.41&0.58&\textbf{0.49}\\
     $\text{Spoken}_{l_{\mathrm{sbert}085}}$ &0.20&0.24&0.56&0.27&813.92&7.34&4316&0.99&0.42&0.59&\textbf{0.49}\\
     $\text{Spoken}_{l_{\mathrm{sbert}090}}$ &0.20&0.22&0.58&0.21&609.18&5.65&4007&0.99&0.43&0.60&\textbf{0.49}\\
     $\text{Spoken}_{l_{\mathrm{sbert}095}}$ &0.18&0.21&0.61&0.16&419.20&4.12&3532&0.99&0.44&0.61&\textbf{0.49}\\
     $\text{Spoken}_{l_{\mathrm{literal}}}$ &0.14&0.18&0.67&0.04&93.74&0.99&2340&1.00&0.46&0.62&0.50\\
  \hline
\end{tabular}
\end{table*}

\section{Results}
The machine evaluation results are shown in Tables 1 and 2. Multi, JMulti, and Spoken refer to the evaluation datasets MultiWOZ, JMultiWOZ, and SpokenWOZ, respectively. The strings sbert* and literal represent the labels used to train PCMs.

From Table 1, we can see that the P-Gain of PCMs fine-tuned with $l_{\mathrm{sbert}T}$ is about twice that of the PCM fine-tuned with $l_{\mathrm{literal}}$. Table 2 shows that the prediction gain of the proposed PCMs exceeds 400 ms, even in the case of $T = 0.95$. The PR$<$R suggests that the response groups prefetched by the PCM trained with an appropriate semantic similarity threshold $T$ cannot be distinguished from the actual response groups. When $T = \{0.90,0.95\}$, PR$<$R is less than $0.50$ for the PCM trained on MultiWOZ, and when $T = 0.95$, PR$<$R is less than $0.50$ for the PCM trained on JMultiWOZ. On the other hand, the semantic similarity and ROUGE evaluations show that the responses deteriorate as $T$ decreases. Looking at the range of $T$ where PR$<$R is less than $0.50$, the results suggested that Japanese requires a higher semantic similarity threshold than English to maintain the quality of the prefetched response group. The difference is probably due to the syntactic nature of Japanese, where the end of a sentence is likely to be the head.
\begin{figure}[t]
\begin{minipage}[b]{1\columnwidth}
    \centering
    \includegraphics[width=0.75\columnwidth]{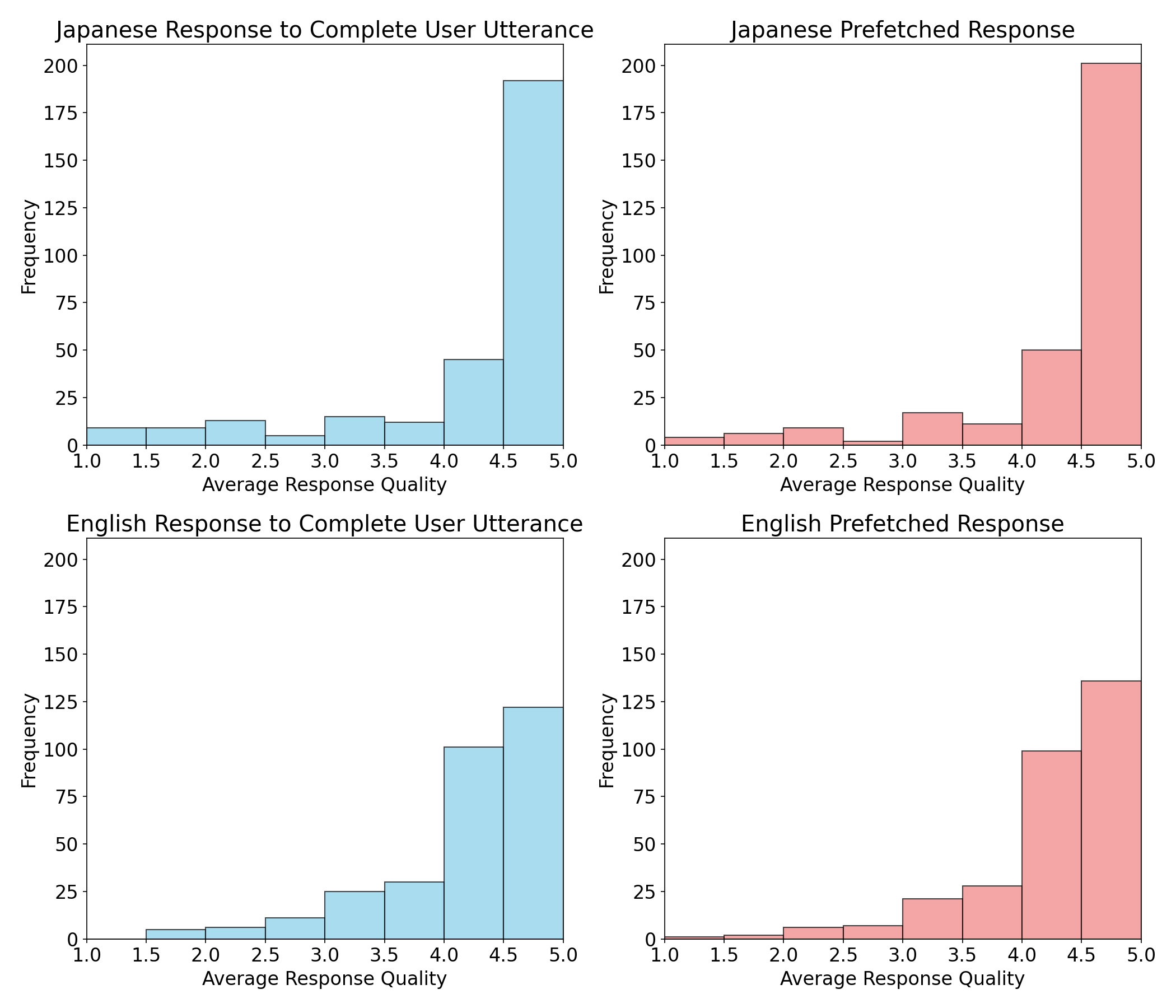}  
    \caption[]{5-point scale evaluation on response naturalness}
    \label{fig:2}
\end{minipage}
\end{figure}

Figure 2 and Table 3 present the human response evaluation results of $\text{Multi}_{l_{\mathrm{sbert}090}}$ and $\text{JMulti}_{l_{\mathrm{sbert}095}}$, where PR$<$R was smaller than 0.50 for the first time at the minimum $T$. For English response naturalness, the prefetched response had an average score $\mu$: $4.17 $ (variance $\sigma^2$: $0.54$), and the actual response had $\mu = 4.25$ ($\sigma^2 = 0.49$). For Japanese response naturalness, the prefetched response had $\mu =  4.28$ ($\sigma^2 = 1.08$), and the actual response had $\mu = 4.42$ ($\sigma^2 = 0.76$). The Pearson correlation between the actual and prefetched response histograms with 8 bins was 0.996 for English and 0.999 for Japanese. It was shown that the English PCM could prefetch responses that humans would perceive as natural if $T$ was set to 0.95 or higher, while the Japanese PCM achieved similar results when $T$ was set to 0.90 or higher. Table 3 shows that in the comparison evaluations of both English and Japanese responses, PR$<$R is below 0.50, suggesting that humans are not able to distinguish between the prefetched responses and actual responses.

\begin{table}[t]
\centering
\caption{Comparison Evaluation on response naturalness}
\begin{tabular}{llll}
\hline
Model & PR$<$R & PR$>$R & PR$=$R \\ \hline
$\text{Multi}_{l_{\mathrm{sbert}090}}$ & 0.20  & 0.30   & 0.50  \\ 
$\text{JMulti}_{l_{\mathrm{sbert}095}}$ & 0.08 & 0.11 & 0.81   \\ \hline
\end{tabular}
\end{table}

\section{Discussion and Conclusion}
We proposed a PCM to estimate the probability of the semantic similarity between the predicted utterance and the complete utterance. As a result, we showed that our PCM greatly reduces user-perceived latency while also maintaining the quality of the system's prefetched response. However, in some settings, the response to a prefetched utterance may be poorer than the response to a full user utterance. The decision to prefetch dynamically for these examples is a topic for future work. Our experiment also shows language-dependent differences in the appropriate semantic similarity threshold for our PCM. The Japanese PCM required a higher semantic similarity threshold to maintain natural response quality compared to the English PCM.

Although our PCM achieves a higher prediction gain than existing PCMs under ideal conditions, a more realistic analysis, considering non-ideal factors such as ASR errors, and response generation time, is necessary for practical applications.

\clearpage

%\section{Acknowledgement}
%Part of this research was supported by the JST Moonshot %Research and Development Program JPMJMS2236.
\section{Acknowledgement}
A part of this work was supported by JST MOONSHOT Grant Number JPMJMS2236 and JSPS KAKENHI 22K17958.
\bibliographystyle{IEEEtran}
\bibliography{mybib}

\end{document}